\pdfoutput=1

\documentclass{article}
\PassOptionsToPackage{numbers, compress}{natbib}
\usepackage[preprint]{neurips_2025}

\usepackage{array}
\usepackage{makecell}

\usepackage{url}
\usepackage{booktabs}
\usepackage{hyperref}
\usepackage{graphicx}
\usepackage{microtype}
\usepackage{multicol}
\usepackage{xcolor}   
\usepackage{multirow}
\usepackage{lipsum}
\usepackage{enumitem}
\usepackage{pifont}
\usepackage{colortbl}
\usepackage{tabularx}
\usepackage{caption} 
\usepackage{float}
\usepackage{amsmath,amsfonts,amssymb,bbm}

\title{\textsc{VLCache}: Computing 2\% Vision Tokens and Reusing 98\% for Vision–Language Inference}

\author{%
  \textbf{Shengling Qin}\textsuperscript{1}\thanks{Equal contribution.} \quad
  \textbf{Hao Yu}\textsuperscript{1}\footnotemark[1] \quad
  \textbf{Chenxin Wu}\textsuperscript{2}\footnotemark[1] \quad
  \textbf{Zheng Li}\textsuperscript{1} \\
  \textbf{Yizhong Cao}\textsuperscript{1} \quad
  \textbf{Zhengyang Zhuge}\textsuperscript{1} \quad
  \textbf{Yuxin Zhou}\textsuperscript{1} \quad
  \textbf{Wentao Yao}\textsuperscript{1} \quad
  \textbf{Yi Zhang}\textsuperscript{1} \\
  \textbf{Zhengheng Wang}\textsuperscript{2} \quad
  \textbf{Shuai Bai}\textsuperscript{1} \quad
  \textbf{Jianwei Zhang}\textsuperscript{1}\thanks{Corresponding author.} \quad
  \textbf{Junyang Lin}\textsuperscript{1} \\
  \\
  \textsuperscript{1}Qwen Team, Alibaba Inc. \\
  \textsuperscript{2}TairKVCache Team, Alibaba Cloud
}

\begin{document}
\maketitle

\begin{abstract}

This paper presents VLCache, a cache reuse framework that exploits both Key-Value (KV) cache and encoder cache from prior multimodal inputs to eliminate costly recomputation when the same multimodal inputs recur.
Unlike previous heuristic approaches, we formally identify the cumulative reuse error effect and demonstrate how to minimize the non-prefix cache reuse error effectively. We further analyze the varying importance of model layers and propose a dynamic, layer-aware recomputation strategy to balance accuracy and efficiency.
Experimental results show that VLCache achieves an accuracy on par with full recomputation, while requiring only 2–5\% of the tokens to compute, yielding 1.2x–16x TTFT speedups.
We develop an experimental implementation of the proposed VLCache pipeline based on SGLang, enabling significantly faster inference in practical deployments.

\end{abstract}

\section{Introduction}
\label{sec:intro}

The Vision-Language Model (VLM) has emerged as a fundamental paradigm for visual question answering and multimodal understanding. With a Vision Transformer (ViT)~\citep{dosovitskiy2020vit,Vaswani2017Attention} as the visual embedding backbone, and a Large Language Model (LLM)~\citep{qwen,qwen2,qwen3} as the main reasoning module, the VLM shows remarkable performance in complex tasks, such as visual reasoning, referring expression comprehension, and multimodal dialogue.

To speed up the inference of language models, engineers leverage the Key-Value (KV) cache computed for a previous request in future requests to skip time-consuming computation. This attention mechanism requires an exact prefix match for lossless cache reuse and token generation.
Considering that the prefix match requirement is not always met, several position-independent reuse schemes, including CacheBlend~\citep{Yao2025CacheBlend}, MPIC~\citep{zhao2025MPIC} and KVShare~\citep{Yang2025KVShare}, have emerged to address this problem. Those methods enable the reuse of the KV cache across different requests, even when initial token sequences differ, thereby eliminating the dependency on shared prefixes. To partially correct the reuse error, these algorithms will recompute part of the tokens at designated positions. Then the problem becomes deciding the recomputation rate and the position of recomputed tokens.
 
However, these heuristic schemes show significant issues in defining the \textit{importance} of the vision tokens, which is the basis of selecting the top ones for recomputation.
For CacheBlend and KVShare, the KV cache distance or the attention score distance between the reused and recomputed tokens is used to decide the importance of each token. Although effective on some occasions, this local information at the layer level cannot accurately tell the importance of the token.
We find that, when evaluating the attention outputs, the token with the greatest distance is often not the most important one. For MPIC, the recomputation budget is allocated to several earlier tokens of each reused cache chunk. The effectiveness of MPIC relies on the \textit{attention sink} effect, where the initial tokens in the prompt—typically the system prompt—attract high attention. In our setting, however, this effect does not occur because the reused vision tokens do not function as a system prompt.

To address these issues, we have an in-depth investigation into reuse errors in VLMs. We identify the \textit{cumulative error effect} on VL cache reuse, which propagates the initial error to later tokens. We argue that this cumulative error is a key reason why \textit{recomputing the initial image tokens} can effectively mitigate performance degradation.
In addition, we observe that different transformer layers contribute unevenly to the final model output. Building on this insight, we further analyze the \textit{layer-wise importance diversity} in VLMs. Based on the error cumulative effect we have discovered, we propose a dynamic algorithm that allocates an optimal per-layer recomputation rate given a fixed overall recomputation budget.

Integrating these findings, we propose \textsc{VLCache}, an end-to-end vision token KV cache reuse pipeline. VLCache stores the output of the vision encoder for each image patch along with the corresponding KV cache. When the same image is encountered again in a subsequent request, VLCache bypasses the vision encoder computation entirely and directly reuses the stored encoder cache and KV cache. To preserve output accuracy, VLCache selectively recomputes a layer-dependent fraction of the initial image tokens at each LLM layer. 
Compared with previous methods, extensive experiments demonstrate that VLCache offers two key advantages: higher acceleration and better accuracy. With only 2-5\% of vision tokens recomputed, VLCache achieves near-lossless accuracy. We integrate the VLCache pipeline into the SGLang framework, significantly reducing the \textit{Time-To-First-Token (TTFT)}. Depending on the model architecture and the length of the input image tokens, VLCache yields 1.2x to 16x TTFT speedups.

\section{Preliminaries}

\begin{figure}[tbp]
    \centering
    \includegraphics[width=1.0\linewidth]{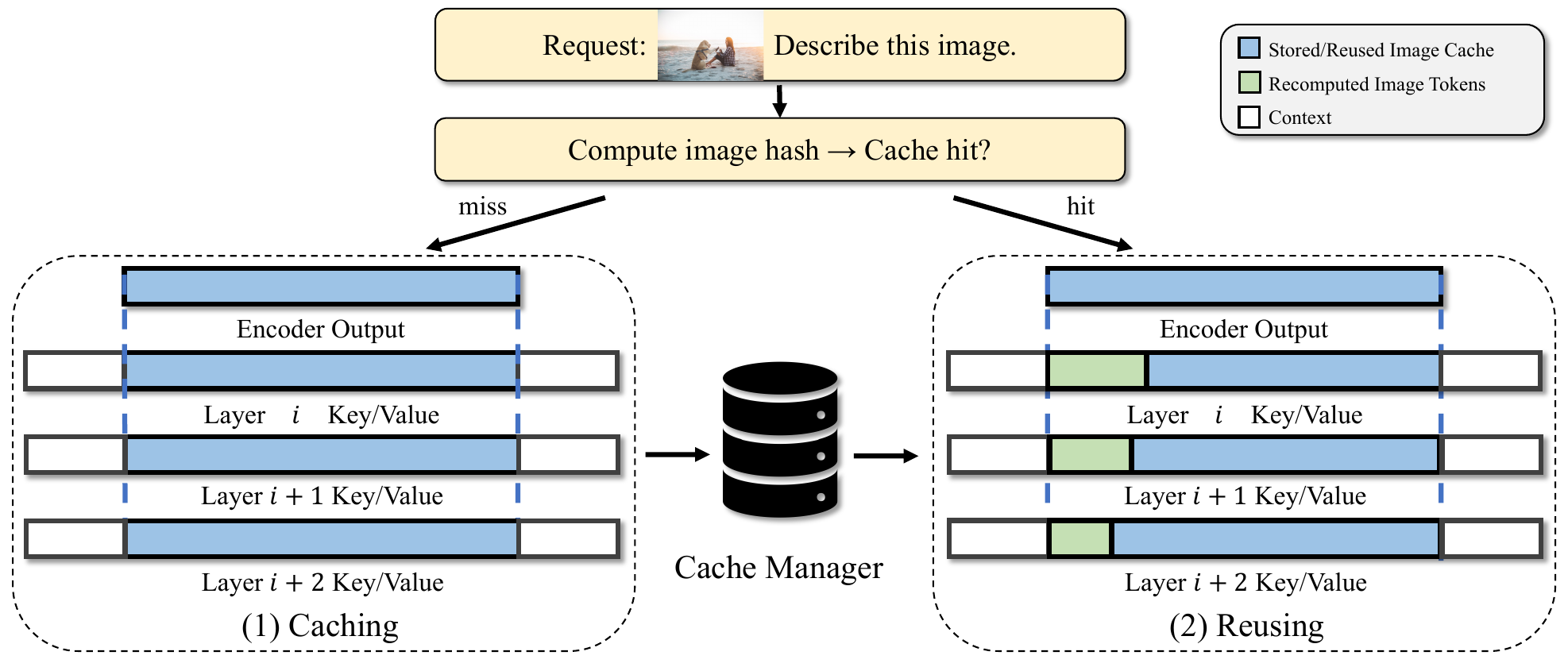}
    \caption{
        Overview of \textsc{VLCache}. Both the KV cache and encoder cache are stored and will be reused in future requests, with selective recomputation in the KV cache to ensure accuracy.
    }
    \label{fig:vlcache}
\end{figure}

\subsection{KV Cache for LLMs}
Vision language models (VLMs)~\citep{Qwen-VL,Qwen2-VL,Qwen2.5-VL} have demonstrated human-level capabilities in understanding and generating multimodal content. State-of-the-art VLMs typically employ a Vision Transformer (ViT)~\citep{Dehghani2023Patch,dosovitskiy2020vit} as the visual encoder, which is then integrated into a large-scale pre-training large language model (LLM) backbone~\citep{qwen, qwen2, qwen2.5}. In general, LLMs are built on the neural structure of Transformers~\citep{Vaswani2017Attention}, which requires a computational cost quadratic to the input sequence's length and makes long sequence inference intractable. To mitigate this inefficiency, the auto-regressive decoding LLMs support Key-Value (KV) caches, i.e., to cache the previous context's intermediate key and value states in memory. KV caches can avoid redundant computation of previous tokens, thus expediting the inference process. Building upon KV cache, the prefix caching~\citep{kwon2023efficient,Zheng2024SGLang} mechanism will retain the KV cache of fixed or shared prefix sequences (such as system prompts or task instructions). This further reduces computation overhead by allowing the model to skip processing the repeated prefix entirely during inference.

\subsection{Position-Independent KV Cache Reuse}
The prefix cache mechanism stipulates that a new request can only reuse KV cached content if its initial segment (i.e., prefix) exactly matches an existing cached sequence. While this strict matching criterion ensures output accuracy, it introduces a significant efficiency bottleneck. Position-independent KV Cache reuse, on the other hand, allows for reusing the pre-cached KV Cache at different locations, thereby achieving higher flexibility. To maintain accuracy after reusing position-independent KV Cache, it is necessary to recompute part of the token's KV Cache. To determine the token indices that need to be recomputed, different researchers have proposed various methods. CacheBlend~\citep{Yao2025CacheBlend} will first calculate the distances of the first three layers of KV Cache, record the farthest token index, and then recompute these tokens in the remaining layers. EPIC~\citep{hu2025epic} and MPIC~\citep{zhao2025MPIC} will recompute the initial tokens of each chunk. Depending on the deviation of the attention map, KVShare~\citep{Yang2025KVShare} will selectively recompute critical tokens during the prefilling and decoding stages. However, all those methods lack both an in-depth analysis of how reused tokens affect the decoding stage and a systematic investigation into the impact of different model layers on the final output.

\section{Observations}
\label{sec:ob}

\subsection{Cumulative Reuse Error Effect}


The self-attention mechanism implies that the output of a token is influenced by its surrounding context. In causal language models, where each token attends to itself and all preceding tokens, the output depends on itself and its prefix. When a multimodal input is reused in a new request, its context differs from the original one—this discrepancy, known as \textit{prefix mismatch}, leads to what we call \textit{reuse error} in the KV cache.
For causal language models, the reuse error of a token consists of two components: the \textit{self reuse error} and the \textit{propagated error} from previously reused tokens.
The propagated error from previous tokens to the current token can then be expressed as:
\begin{align}
    e_{k}^{prop} = \sum_{i=1}^{k-1} e_{(i, k)}^{prop},
\end{align}
where $e_{(i, k)}^{prop}$ represents the propagated error from the reused token at position $i$ to position $k$. Thus, the total reuse error for token $k$ is given by:
\begin{align}
e_{k}^{total} &= e_{k}^{self} + e_{k}^{prop} \\
&= e_{k}^{self} + \sum_{i=1}^{k-1} e_{(i, k)}^{prop} \\
&= e_{k}^{self} +  e_{(1, k)}^{prop} + \cdots +  e_{(k-2, k)}^{prop} + e_{(k-1, k)}^{prop}.
\end{align}

Therefore, the `accuracy' of a reused token is affected by all previously reused tokens, while it, in turn, influences all subsequent tokens. We refer to this phenomenon as \textit{reuse error propagation}. This property introduces positional weighting among tokens and serves as the theoretical foundation for designing an efficient recomputation token selection strategy.

\begin{figure}[t]
    \centering
    \includegraphics[width=1.0\linewidth]{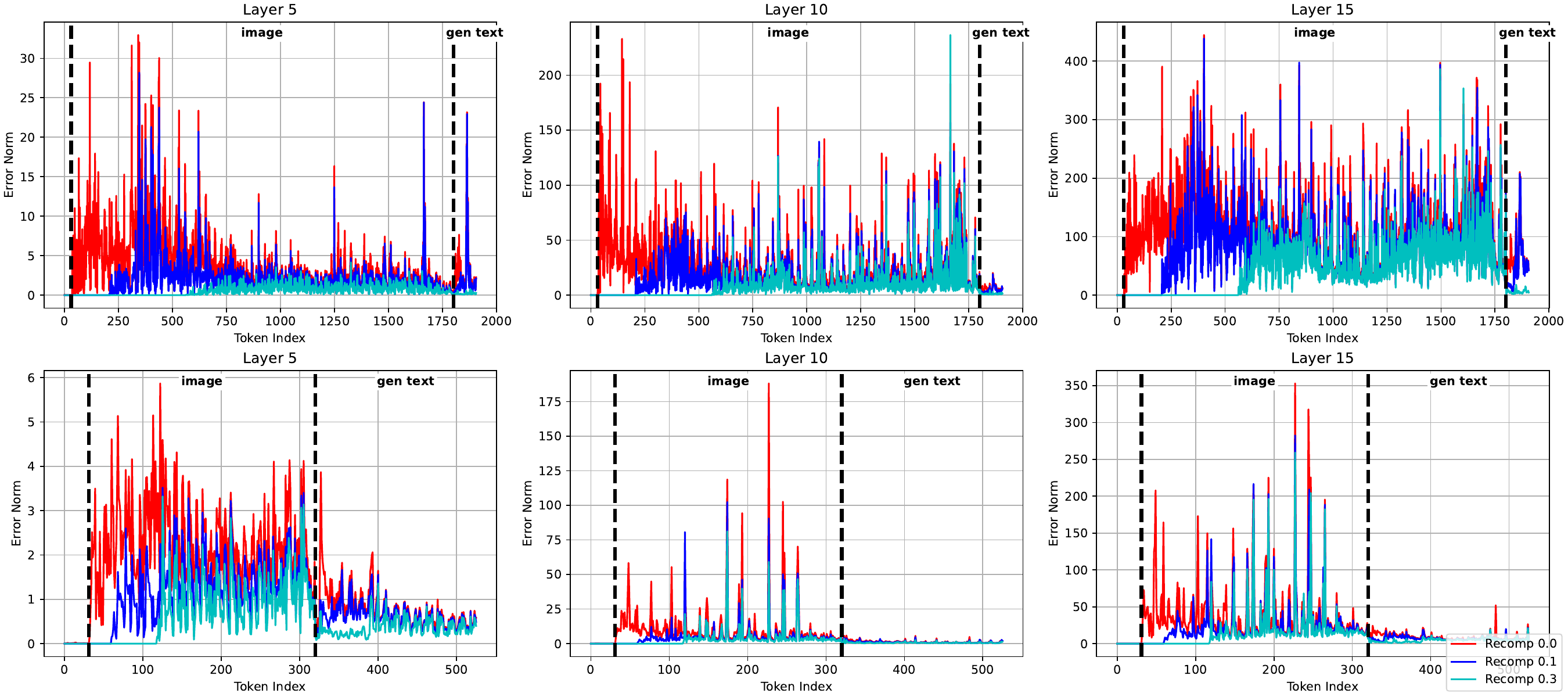}
    \caption{
        The reuse error of earlier token recomputation for two image examples. Earlier token recomputation cancels out more reuse errors and slows down the error accumulation at later generated tokens, leading to better accuracy.
    }
    \label{fig:attn_error}
\end{figure}

To validate the propagation of the reuse error, we conduct several experiments shown in Fig. \ref{fig:attn_error}.
Using two representative image inputs, we measure each token’s attention output error when its key/value states are reused rather than recomputed, and compare per-token error norms under different recomputation ratios throughout the generation sequence.
In both examples, the propagation pattern is clear: early reused tokens introduce errors that persist and accumulate, yielding larger error norms for later tokens. Selectively recomputing early tokens (e.g., 10\% or 30\%) significantly suppresses downstream accumulation, as early recomputation cancels a large portion of the upstream error. These findings directly support our theoretical claim that reuse error induces a positional weighting effect and show that prioritizing earlier tokens for recomputation offers disproportionately large accuracy gains.

\subsection{Layer-wise Importance Difference}
\label{sec:layerwise_importance_difference}
As mentioned above, different model layers exert distinct influences on the output. To validate this hypothesis, we conducted an empirical experiment on Qwen2.5-VL-7B. Specifically, we select 20 samples each from the MMMU-Pro~\citep{yue2025mmmupro} Standard \& Vision dataset used as experimental subjects. The entire experimental procedure consists of the following three phases:
\begin{itemize}
    \item Constructing the KV cache with a replacement prompt. For each sample, we keep the image input unchanged and replace the original question prompt with a semantically unrelated or neutral alternative prompt (e.g., `Please describe this image'). In this setting, we perform a full forward pass and cache the image-related KV cache. 
    \item Obtaining the original output as the baseline. Next, we conduct standard decoding with the original prompt, recording the output logits and corresponding text. 
    \item Reuse KV cache and measure errors. Finally, we combine the original question prompt and the text generated via standard decoding as input. During inference, we inject the image KV cache generated from the substituted prompt (i.e., a `mismatched' KV cache) into all layers, while recomputing the KV cache for only the top 10\% \& 20\% \& 30\% of image tokens in a single layer. We record the logits of the decoding text and compute the MSE loss between them and the logits produced by the original model.
\end{itemize}

\begin{figure}[tbp]
    \centering
    \includegraphics[width=0.44\linewidth]{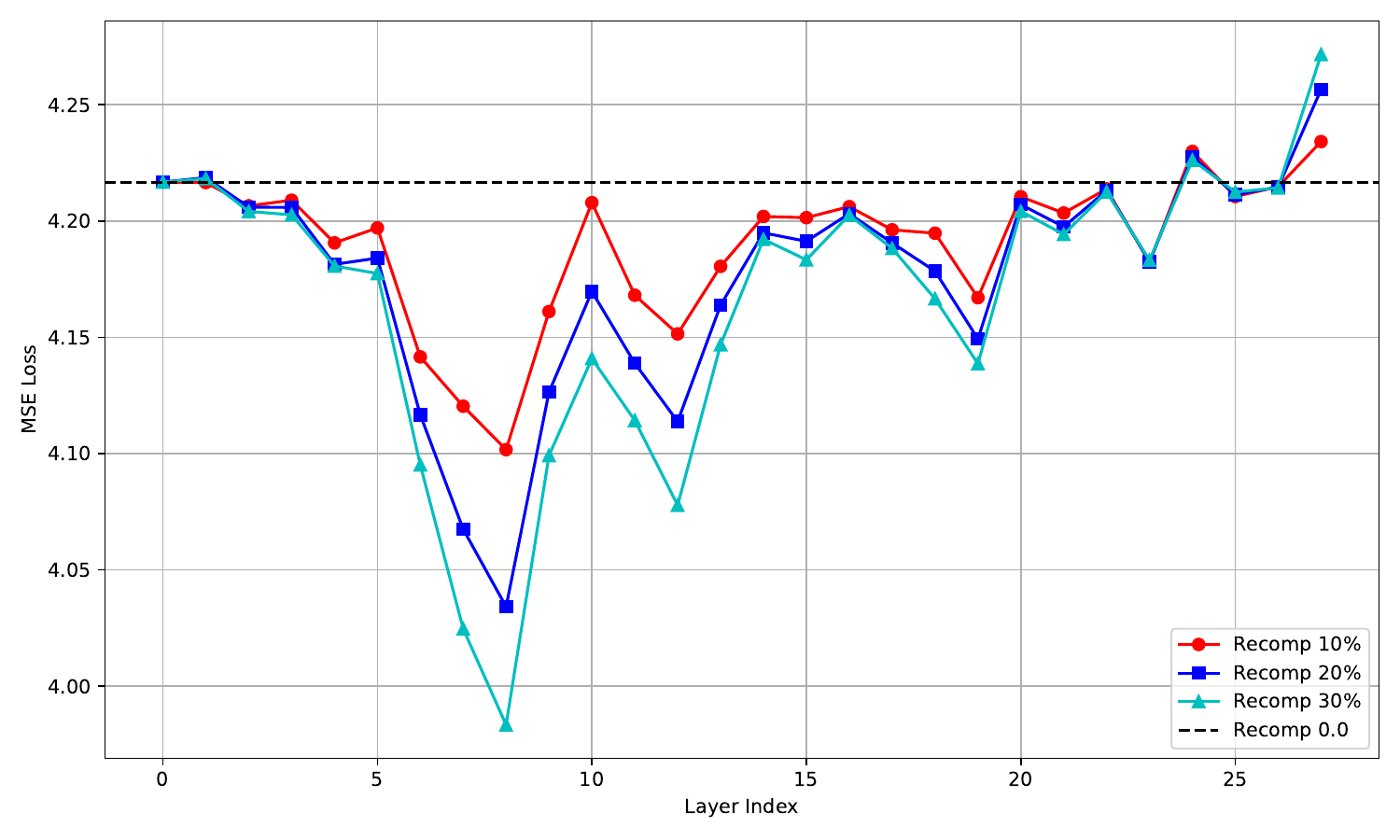}
    \hspace{1em}%
    \includegraphics[width=0.44\linewidth]{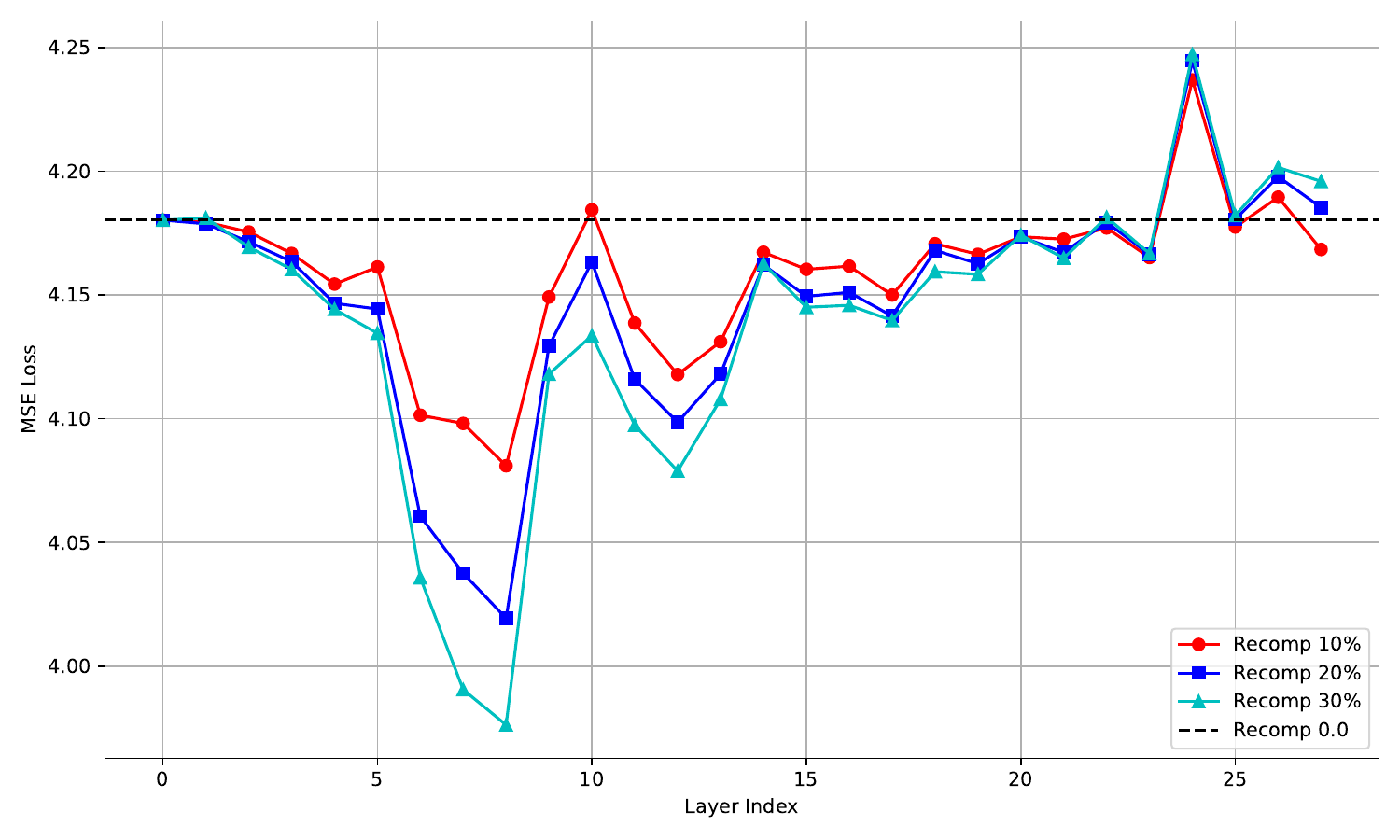}
    \caption{MSE loss between the logits of the generated text and those produced by the original model (with correct KV cache), when recomputing the top 10\%, 20\%, and 30\% of image tokens, evaluated on the MMMU-Pro Standard (left) and MMMU-Pro Vision (right) datasets.}
    \label{fig:layerwise-error}
\end{figure}
The loss curve is shown in Fig~\ref{fig:layerwise-error}. Compared to directly reusing the KV cache, applying different recomputation ratios across layers leads to varying degrees of MSE loss reduction for the decoding text.
These results show that using the same recomputation ratio across all layers is \textit{sub-optimal}, as some more sensitive layers should have a higher recomputation ratio, while some layers should not even recompute image tokens at all.


\section{Algorithm}

\subsection{VLCache: Reuse and Dynamic Recomputation}

\paragraph{Basic Algorithm}

\begin{figure}[t]
    \centering
    \includegraphics[width=0.8\linewidth]{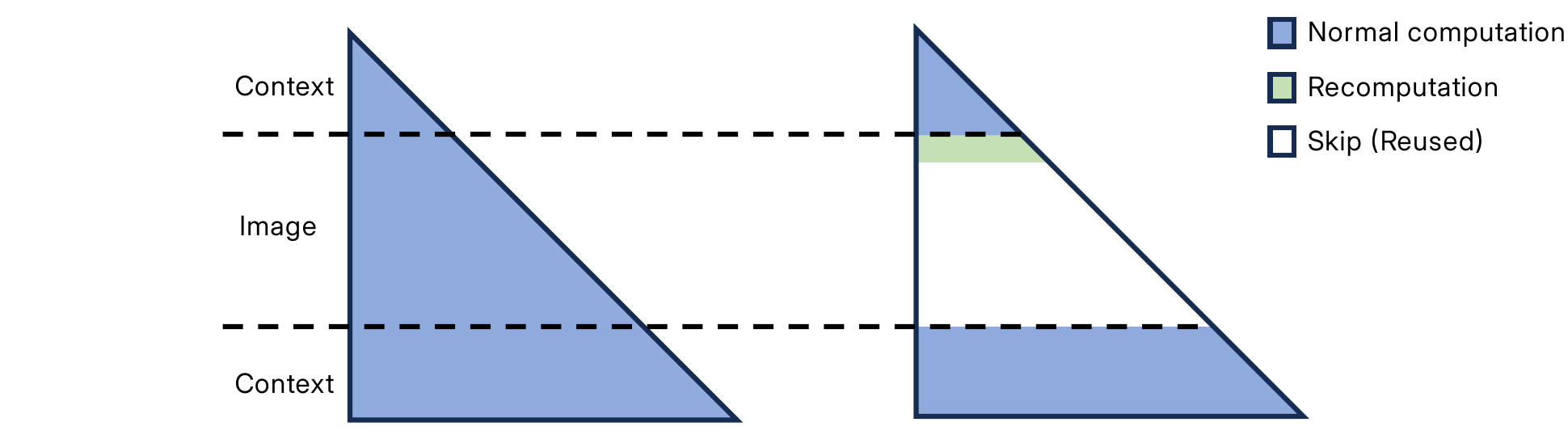}
    \caption{
        Full attention computation (left) v.s. VLCache partial attention computation (right). VLCache skips a large portion of attention operations and forms contiguous computation regions, which is hardware-friendly.
    }
    \label{fig:attn_skip}
\end{figure}

Based on the above observations, we design the VLCache algorithm illustrated in Fig. \ref{fig:vlcache}, which reuses most vision tokens while selectively recomputing a small subset of important ones.
VLCache operates as a plugin within the standard inference pipeline. Upon receiving a request, it first computes a hash of the input image. For cache misses, normal prefilling is performed and both the encoder cache and KV cache are stored in the backend KVCache store.
For cache hits, the stored encoder and KV caches are retrieved and reused.
We directly reuse most image tokens and only recompute a small set of early tokens, with the recomputation ratio denoted as $r$. The query tokens and the selected tokens for recomputation are concatenated to form the actual model input. For reused tokens, all associated computations—including attention and MLP—will be skipped, which determines the acceleration. An example of attention skipping is shown in Fig. \ref{fig:attn_skip}. The resulting attention computation regions become contiguous, making the kernel execution more hardware-friendly.

\paragraph{Dynamic Recomputation}

As previously discussed, a key challenge lies in accurately determining the proportion of tokens to recompute for each layer. To address this, we propose an adaptive search strategy, i.e., to allocate a higher recomputation ratio to layers that are more sensitive to output quality, while applying a more aggressive KV cache reuse policy to less sensitive layers.

Given a proxy dataset $D$, we follow the experimental procedure outlined in Section~\ref{sec:layerwise_importance_difference}. First, we generate the mismatched KV cache (i.e., using a substituted prompt) and reuse it across all layers. Next, for each layer, we evaluate the MSE loss of the decoding text logits corresponding to the original model and recompute a fixed proportion of image tokens in that layer, i.e.,
\begin{equation}
    \mathcal{S}_{i}(r) = \mathbb{E}_{(x, y) \in \mathcal{D}} 
        \mathrm{MSE}\big( {\mathbf{z}^{\text{orig}}, \mathbf{z}^{\text{reuse}}_{i,r} \big)},
\end{equation}
where $\mathcal{S}_{i}(r)$ is the layer $i$ sensitivity score with recomputation rate $r$ (e.g., r=0.002, 0.004, $\cdots $0.300). $\mathbf{z}^{\text{orig}}$ is the logits produced by the original model using correct KV cache. $\mathbf{z}^{\text{reuse}}_{i,r}$ represents the logits where the mismatched key-value cache is reused across all layers, and only in the $i$-th layer the first $r$ tokens will be recomputed.

After obtaining the sensitivity scores for each layer, given a target total recomputation ratio $ P_{\text{tar}}$, we minimize the sum of sensitivity scores across all layers, subject to the constraint that the per-layer recomputation ratio is non-increasing with depth (i.e., monotonically decreasing or constant across layers). That is
\begin{align}
    & \underset{\{r_i\}_{i=1}^L}{\text{minimize}} 
    && \sum_{\ell=1}^{L} \mathcal{S}_{i}(r) \\
    & \text{subject to}
    && \sum_{i=1}^{L} r_i \leq P_{\text{tar}}, \\
    &&& r_1 \geq r_2 \geq \cdots \geq r_L \geq 0, \\
    &&& r_i \in \{0.002, 0.004, \cdots 0.300\}, \quad \forall i \in \{1, \dots, L\}.
\end{align}

Given a set of distinct target parameters, we can dynamically generate multiple choices, demonstrating remarkable flexibility.

Since the selected ranks are discrete, this optimization problem is inherently an integer programming problem. To address it, we propose a simple greedy algorithm for an approximate solution. Notably, we adopt a simplifying assumption: the sensitivity of a given layer depends only on its own recomputation strategy and is independent of the ranks chosen for other layers. Although this greedy algorithm does not guarantee a globally optimal configuration, exhaustively searching over all possible combinations to find the exact global minimum is computationally prohibitive. In contrast, our approximate greedy approach dramatically accelerates the search process. As we will demonstrate in subsequent experiments, the adaptive configurations obtained via this greedy strategy achieve strong performance across diverse model architectures and a wide range of tasks.

Notably, to ensure the validity of acceleration, the token recomputation rate must be monotonically non-increasing across layers, i.e., $r_i \leq r_{i-1}$  for all layers. This constraint arises from VLCache’s core mechanism: it reuses hidden states of retained tokens from upper layers and skips their corresponding attention and MLP computations in subsequent layers. If a deeper layer were assigned a higher recomputation rate than the shallow layer, it would require recomputing hidden states for tokens that were not retained (and thus not available for reuse) by the shallow layer. This will violate the reuse assumption and introduce redundant computation, which directly contradicts the acceleration objective.

\subsection{SGLang Integration of VLCache}
For evaluation, we implement the proposed VLCache pipeline as a prototype based on the SGLang framework. Our integration introduces two key modifications: (1) skipping ViT computation via hash-based image embedding caching, and (2) skipping attention and MLP computation through per-image adaptive recomputation with fine-grained KV cache reuse.

For the ViT encoder, we implement request-level embedding reuse based on content hashing. Upon receiving a multimodal request, SGLang computes a global hash over the concatenated pixel values of all input images. If this hash matches a previously processed request, the precomputed image embeddings are retrieved from a distributed key-value store which is Tair KVCache Store in practice, entirely bypassing ViT computation. Otherwise, embeddings are computed and stored for future reuse.

To support image-level KV cache reuse, each image in a request is assigned an individual hash based on its pixel content. These per-image hashes enable fine-grained identification of reused visual content across varying contexts, enabling modular cache management.
We store the KV cache before rotary position embedding (RoPE) is applied.
Building on this, we implement adaptive recomputation in different attention layers using \textit{Block Sparse Attention}. At each layer, a recompute ratio $r$ determines the fraction of tokens to recompute per image. For an image with T tokens, only the first $\lfloor r \cdot T \rfloor$ tokens undergo full attention computation; the remainder reuse their KV states from prior executions, indexed by the corresponding image hash. In practice, we construct a \textit{computation mask} at each layer to identify which tokens require computation or not. This binary mask is used to index into the input hidden states, extracting only the subset of tokens designated for fresh processing. The selected tokens are then passed through RoPE and fed into the attention kernel for computation.

These features are seamlessly integrated into SGLang’s execution engine. The system transparently accelerates VLM inference, enabling efficient deployment of long-context multimodal workloads.

\section{Experiments and Discussion}

\subsection{Acceleration of VLCache}
We evaluate the effectiveness of VLCache in reducing \textit{Time to First Token (TTFT)} on Qwen3-VL-8B and Qwen3-VL-32B using an NVIDIA H20-3e GPU. Each input prompt follows the format:
\begin{center}
\verb|<text><image_1><image_2>...<image_n>|,
\end{center}
where the number of images varies across requests, simulating realistic multimodal workloads.
Experiments are conducted with varying total input lengths (including both text and image tokens) under the following four configurations:

\begin{itemize}
\item Origin: Baseline setup with full computation;
\item w/o ViT: Skip ViT computation by reusing cached image embeddings;
\item Static Recompute: Skip ViT using cached image embeddings, and skip attention and MLP computation with a uniform recompute ratio~$r$ applied across all layers;
\item Dynamic Recompute: Skip ViT using cached image embeddings, and skip attention and MLP computation with layer-adaptive recompute ratios. The average recompute ratio is denoted as $\bar{r}$.
\end{itemize}

To measure the sensitivity of a specific layer, we randomly select 40 samples from the training set to construct an auxiliary dataset $D$. The specific recomputation ratio for each model can be viewed in our code.

\begin{table}[t]
\centering
\caption{Speedup for Qwen3-VL-8B}
\label{tab:qwen3_vl_8b_ttft}
\resizebox{\textwidth}{!}{
\begin{tabular}{lcccccccccc}
\toprule
        & \multicolumn{2}{c}{\textbf{1K}} & \multicolumn{2}{c}{\textbf{3K}} & \multicolumn{2}{c}{\textbf{5K}} & \multicolumn{2}{c}{\textbf{10K}} & \multicolumn{2}{c}{\textbf{20K}} \\
        \cmidrule(lr){2-3}\cmidrule(lr){4-5}\cmidrule(lr){6-7}\cmidrule(lr){8-9}\cmidrule(lr){10-11}
        \textbf{Configuration} & \textbf{TTFT} & \textbf{Speedup} & \textbf{TTFT} & \textbf{Speedup} & \textbf{TTFT} & \textbf{Speedup} & \textbf{TTFT} & \textbf{Speedup} & \textbf{TTFT} & \textbf{Speedup} \\
\midrule
        Origin                     & 0.392 & 1.00x & 1.176 & 1.00x & 1.993 & 1.00x & 4.087 & 1.00x & 8.784 & 1.00x \\
        w/o ViT                    & 0.286 & 1.37x & 0.962 & 1.22x & 1.631 & 1.22x & 3.428 & 1.19x & 7.565 & 1.16x \\
        Static (r=0.3)             & 0.249 & 1.57x & 0.794 & 1.48x & 1.306 & 1.53x & 2.592 & 1.58x & 5.341 & 1.64x \\
        Static (r=0.2)             & 0.229 & 1.71x & 0.752 & 1.56x & 1.235 & 1.61x & 2.428 & 1.68x & 4.945 & 1.78x \\
        Static (r=0.1)             & 0.230 & 1.70x & 0.720 & 1.63x & 1.208 & 1.65x & 2.361 & 1.73x & 4.666 & 1.88x \\
        Static (r=0.0)             & 0.227 & 1.73x & 0.692 & 1.70x & 1.133 & 1.76x & 2.206 & 1.85x & 4.430 & 1.98x \\
        Dynamic ($\bar{r}=0.035$)  & 0.253 & 1.55x & 0.750 & 1.57x & 1.222 & 1.63x & 2.408 & 1.70x & 4.823 & 1.82x \\
        Dynamic ($\bar{r}=0.025$)  & 0.259 & 1.51x & 0.728 & 1.62x & 1.258 & 1.58x & 2.438 & 1.68x & 4.714 & 1.86x \\
\bottomrule
\end{tabular}
}
\end{table}

\begin{table}[t]
\centering
\caption{Speedup for Qwen3-VL-32B}
\label{tab:qwen3_vl_32b_ttft}
\resizebox{\textwidth}{!}{
\begin{tabular}{lcccccccccc}
\toprule
        & \multicolumn{2}{c}{\textbf{1K}} & \multicolumn{2}{c}{\textbf{3K}} & \multicolumn{2}{c}{\textbf{5K}} & \multicolumn{2}{c}{\textbf{10K}} & \multicolumn{2}{c}{\textbf{20K}} \\
        \cmidrule(lr){2-3}\cmidrule(lr){4-5}\cmidrule(lr){6-7}\cmidrule(lr){8-9}\cmidrule(lr){10-11}
        \textbf{Configuration} & \textbf{TTFT} & \textbf{Speedup} & \textbf{TTFT} & \textbf{Speedup} & \textbf{TTFT} & \textbf{Speedup} & \textbf{TTFT} & \textbf{Speedup} & \textbf{TTFT} & \textbf{Speedup} \\
\midrule
        Origin                     & 0.772 & 1.00x & 2.501 & 1.00x & 4.186 & 1.00x & 8.927 & 1.00x & 20.279 & 1.00x \\
        w/o ViT                    & 0.701 & 1.10x & 2.218 & 1.13x & 3.761 & 1.11x & 8.078 & 1.11x & 18.549 & 1.09x \\
        Static (r=0.3)             & 0.423 & 1.83x & 1.337 & 1.87x & 2.183 & 1.92x & 4.335 & 2.06x & 9.102 & 2.23x \\
        Static (r=0.2)             & 0.368 & 2.10x & 1.098 & 2.28x & 1.901 & 2.20x & 3.787 & 2.36x & 7.794 & 2.60x \\
        Static (r=0.1)             & 0.330 & 2.34x & 0.963 & 2.60x & 1.551 & 2.70x & 2.984 & 2.99x & 6.099 & 3.32x \\
        Static (r=0.0)             & 0.306 & 2.52x & 0.792 & 3.16x & 1.313 & 3.19x & 2.522 & 3.54x & 5.144 & 3.94x \\
        Dynamic ($\bar{r}=0.035$)  & 0.393 & 1.96x & 0.995 & 2.51x & 1.643 & 2.55x & 3.136 & 2.85x & 6.399 & 3.17x \\
        Dynamic ($\bar{r}=0.025$)  & 0.383 & 2.02x & 1.023 & 2.44x & 1.651 & 2.54x & 3.160 & 2.83x & 6.375 & 3.18x \\
\bottomrule
\end{tabular}
}
\end{table}

Table \ref{tab:qwen3_vl_8b_ttft} - \ref{tab:qwen3_vl_32b_ttft} present the detailed TTFT acceleration results. Here, the text remains fixed at 20 tokens, and the terms like `1K', `3K' refer to the length of the image tokens. For Qwen3-VL series models, we observe substantial speedups of up to 3.9x. The acceleration is predominantly attributable to KV cache reuse, as a result of the more efficient ViT module architecture in the model.
The speedup becomes more pronounced as the number of image tokens increases, since KV cache reuse effectively reduces the computationally intensive attention operation with $O(n^2)$ complexity.
To minimize the recomputation rate, we employ a dynamic recomputation scheme, which introduces a small overhead but maintains significant speedup, especially for longer input sequences.

\subsection{Accuracy Preservation of VLCache}

We assess the near-lossless performance of VLCache across a broad suite of vision–language benchmarks, covering general visual question answering, visual mathematics, OCR-based tasks, and hallucinations.
The experiments are conducted in the following manner:
\begin{itemize}
    \item Extract the images from the dataset, and pair each image with a relevant query to construct an auxiliary prompt;
    \item Construct the encoder cache and KV cache pool by input the auxiliary prompt;
    \item Send the formal prompt and evaluate the model response.
\end{itemize}


\begin{table}[t]
\centering
\caption{Accuracy Preservation for Qwen3-VL-8B}
\label{tab:acc_3_8b}
\resizebox{\textwidth}{!}{
\begin{tabular}{lccccccc}
\toprule
\textbf{Dataset} &
$r=0.0$ &
$\bar{r}=0.025$ &
$\bar{r}=0.035$ &
$r=0.1$ &
$r=0.2$ &
$r=0.3$ &
$r=1.0$ \\
\midrule
MMMU-VAL            & 67.44 & 67.33 & 67.00 & \textbf{68.00} & 67.22 & 66.56 & 66.33 \\
MMMU-Pro-Standard   & 53.82 & 54.28 & \textbf{54.68} & 54.05 & 53.53 & 54.51 & 54.10 \\
MMMU-Pro-Vision     & 53.12 & 53.29 & \textbf{53.93} & 52.49 & 53.29 & 52.60 & 52.14 \\
MathVista-MINI      & 77.00 & 77.20 & 77.00 & 77.60 & 77.60 & 77.70 & \textbf{77.80} \\
HallusionBench      & 58.97 & 60.17 & 60.28 & 60.45 & 59.18 & 60.59 & \textbf{61.78} \\
MBench-DEV-CN-V11   & 84.13 & 84.37 & 84.29 & 84.06 & 84.37 & 84.75 & \textbf{85.53} \\
MBench-DEV-EN-V11   & 85.14 & 85.91 & 85.84 & 85.45 & 85.91 & \textbf{86.15} & 85.91 \\
RealWorldQA         & 71.13 & 71.76 & 71.76 & 71.76 & \textbf{72.03} & 71.37 & 70.98 \\
AI2D-TEST           & 85.27 & 84.75 & 85.01 & 85.20 & 84.68 & \textbf{85.49} & 85.43 \\
ChartQA-TEST        & 83.60 & \textbf{84.44} & \textbf{84.44} & \textbf{84.44} & 84.08 & 84.12 & 84.24 \\
DocVQA-VAL          & 94.84 & \textbf{95.15} & \textbf{95.15} & 95.08 & 95.13 & 95.14 & 95.09 \\
\midrule
\textbf{Mean}        & 74.04 & 74.42 & \textbf{74.49} & 74.42 & 74.27 & 74.45 & 74.48 \\
\bottomrule
\end{tabular}
}
\end{table}

\begin{table}[t]
\centering
\caption{Accuracy Preservation for Qwen3-VL-32B}
\label{tab:acc_3_32b}
\resizebox{\textwidth}{!}{
\begin{tabular}{lccccccc}
\toprule
\textbf{Dataset} &
$r=0.0$ &
$\bar{r}=0.025$ &
$\bar{r}=0.035$ &
$r=0.1$ &
$r=0.2$ &
$r=0.3$ &
$r=1.0$ \\
\midrule
MMMU-VAL            & 73.78 & 74.44 & 73.44 & 74.11 & 73.67 & \textbf{74.78} & 73.00 \\
MMMU-Pro-Standard   & 64.57 & 63.93 & \textbf{65.20} & 63.76 & 65.09 & 63.70 & 63.18 \\
MMMU-Pro-Vision     & 61.01 & 61.21 & \textbf{62.77} & 62.60 & 62.02 & 62.43 & 61.68 \\
MathVista-MINI      & 82.90 & 82.90 & 82.90 & \textbf{83.30} & 82.70 & 83.00 & 82.10 \\
HallusionBench      & 62.53 & 63.52 & 62.90 & 62.43 & 61.81 & 63.48 & \textbf{64.13} \\
MBench-DEV-CN-V11   & 88.70 & 89.01 & \textbf{89.40} & 89.09 & 89.01 & 88.62 & 88.70 \\
MBench-DEV-EN-V11   & 89.32 & 89.09 & 89.09 & 89.32 & 89.09 & \textbf{89.55} & 89.47 \\
RealWorldQA         & 79.35 & \textbf{80.65} & 79.87 & 80.26 & 79.08 & 78.95 & 78.69 \\
AI2D-TEST           & 88.21 & 88.70 & 88.83 & \textbf{88.99} & \textbf{88.99} & 88.80 & 88.89 \\
ChartQA-TEST        & 83.00 & 83.64 & 83.56 & 83.96 & 84.48 & 84.32 & \textbf{86.92} \\
DocVQA-VAL          & 95.83 & 95.78 & 95.84 & 95.80 & \textbf{95.87} & 95.70 & \textbf{95.87} \\
\midrule
\textbf{Mean}        & 79.01 & 79.35 & \textbf{79.44} & 79.42 & 79.26 & 79.39 & 79.33 \\
\bottomrule
\end{tabular}
}
\end{table}

The results are shown in Table \ref{tab:acc_3_8b} - Table \ref{tab:acc_3_32b}. For example, across the datasets in Table~\ref{tab:acc_3_8b}, VLCache maintains accuracy extremely well even at low recomputation rates. For example, at $r = 0.1$, most datasets show accuracy values nearly identical to the $r = 1.0$ (full-recompute) baseline. Datasets such as MMMU-VAL, MathVista-MINI, RealWorldQA, ChartQA-TEST, and DocVQA-VAL all remain within roughly 0–1 percentage points of their full-recompute accuracy. More challenging benchmarks, such as HallusionBench or MBench-DEV-CN-V11, display similarly tight accuracy preservation, with changes typically under 1–2 points.

Besides, our dynamic KV-cache recomputation strategy achieves remarkable accuracy preservation across diverse multimodal benchmarks—even at very low recomputation rates. Notably, the mean accuracy peaks at $\bar{r} =0.035~(74.49)$.
This advantage stems from our layer-wise dynamic allocation of recomputation budget: rather than uniformly recompute tokens across layers (as in static schemes), our method preferentially targets layers and tokens where cache staleness most harms model output—yielding higher fidelity at lower cost. 

Overall, the tables demonstrate that VLCache can maintain 99\% accuracy at low recomputation rates, with accuracy steadily increasing as $r$ grows, and that dynamic recomputation provides better performance compared to static approaches.

\begin{table}[t]
\centering
\caption{Performance comparison of our method and prior SOTA on Qwen3-VL-8B. `Cache' uses KV-cache distance for matching (CacheBlend), while `Attn.' uses attention-map–based matching (KVShare).}
\label{tab:compare_acc_3_8b}
\resizebox{\textwidth}{!}{
\begin{tabular}{lccccccccc}
\toprule
\multirow{2}{*}{Metric} &
\multicolumn{3}{c}{$r=0.1$} &
\multicolumn{3}{c}{$r=0.2$} &
\multicolumn{3}{c}{$r=0.3$} \\
\cmidrule(lr){2-4} \cmidrule(lr){5-7} \cmidrule(lr){8-10}
& Ours & Cache & Attn. & Ours & Cache & Attn. & Ours & Cache & Attn. \\
\midrule
MMMU-VAL           & 68.00 & \textbf{68.89} & 68.45 & 67.22 & \textbf{67.56} & 67.20 & \textbf{66.56} & 65.78 & 65.95 \\
MMMU-Pro-Standard  & 54.05 & \textbf{54.45} & 54.22 & \textbf{53.53} & 53.29 & 53.30 & 54.51 & 54.45 & \textbf{55.08} \\
MMMU-Pro-Vision    & 52.49 & \textbf{53.41} & 52.10 & \textbf{53.29} & 52.31 & 52.65 & 52.60 & \textbf{53.70} & 52.02 \\
MathVista-MINI     & \textbf{77.60} & 76.40 & 75.95 & \textbf{77.60} & 76.70 & 77.15 & \textbf{77.70} & 76.90 & 76.08 \\
MathVision         & \textbf{60.45} & 58.72 & 58.20 & 59.18 & \textbf{60.09} & 59.70 & \textbf{60.59} & 59.29 & 58.40 \\
MBench-DEV-CN-V11  & 84.06 & \textbf{84.13} & 83.55 & 84.37 & \textbf{84.91} & 84.10 & \textbf{84.75} & 84.52 & 83.95 \\
MBench-DEV-EN-V11  & 85.45 & 84.91 & 84.30 & \textbf{85.91} & 85.45 & 85.85 & \textbf{86.15} & 85.72 & 85.10 \\
RealWorldQA        & \textbf{71.76} & 71.37 & 70.80 & \textbf{72.03} & 70.98 & 71.10 & 71.37 & \textbf{71.50} & 70.65 \\
AI2D-TEST          & 85.20 & \textbf{85.98} & 84.85 & 84.68 & \textbf{85.10} & 84.20 & \textbf{85.49} & 85.17 & 84.45 \\
ChartQA-TEST       & \textbf{84.44} & 83.64 & 83.10 & \textbf{84.08} & 83.44 & 82.95 & \textbf{84.12} & 83.44 & 83.80 \\
DocVQA-VAL         & 95.08 & \textbf{95.17} & 94.55 & 95.13 & \textbf{95.19} & 94.60 & 95.14 & \textbf{95.18} & 94.50 \\
\midrule
\textbf{Mean}        & \textbf{74.42} & 74.28 & 73.64 & \textbf{74.27} & 74.09 & 73.89 & \textbf{74.45} & 74.15 & 73.63 \\
\bottomrule
\end{tabular}
}
\end{table}

\subsection{Comparisons with Other SOTA Methods}
We conduct comparison experiments between our proposed VLCache and two state-of-the-art baselines, i.e., CacheBlend~\citep{Yao2025CacheBlend} and KVShare~\citep{Yang2025KVShare} under identical hyperparameter settings. Without employing our dynamic strategy, we adopt a uniform recomputation rate across all layers—specifically 10\%, 20\%, and 30\%.
As shown in Table~\ref{tab:compare_acc_3_8b}, VLCache consistently outperforms both baselines at every recomputation rate, demonstrating its effectiveness, robustness, and the validity of the underlying \textit{cumulative reuse error effect}.

\section{Conclusion}

In this work, we propose VLCache, a cache reuse scheme designed to accelerate the prefill stage of VLMs while preserving inference accuracy.
VLCache provides a holistic design that integrates algorithmic optimizations and system-level implementations, achieving significant acceleration without compromising accuracy.
It demonstrates substantial speedup over standard VLM inference, especially in scenarios where multiple queries are paired with a same set of images to become overlapping requests.
Following this work, future explorations could investigate extending cache reuse to similar visual inputs, enabling broader applicability across diverse multi-query or multi-image tasks.

\newpage
\bibliographystyle{plainnat}
\bibliography{reference}

@article{qwen,
  title={Qwen Technical Report},
  author={Jinze Bai and Shuai Bai and Yunfei Chu and Zeyu Cui and Kai Dang and Xiaodong Deng and Yang Fan and Wenbin Ge and Yu Han and Fei Huang and Binyuan Hui and Luo Ji and Mei Li and Junyang Lin and Runji Lin and Dayiheng Liu and Gao Liu and Chengqiang Lu and Keming Lu and Jianxin Ma and Rui Men and Xingzhang Ren and Xuancheng Ren and Chuanqi Tan and Sinan Tan and Jianhong Tu and Peng Wang and Shijie Wang and Wei Wang and Shengguang Wu and Benfeng Xu and Jin Xu and An Yang and Hao Yang and Jian Yang and Shusheng Yang and Yang Yao and Bowen Yu and Hongyi Yuan and Zheng Yuan and Jianwei Zhang and Xingxuan Zhang and Yichang Zhang and Zhenru Zhang and Chang Zhou and Jingren Zhou and Xiaohuan Zhou and Tianhang Zhu},
  journal={arXiv preprint arXiv:2309.16609},
  year={2023}
}

@article{qwen3,
    title={Qwen3 Technical Report}, 
    author={An Yang and Anfeng Li and Baosong Yang and Beichen Zhang and Binyuan Hui and Bo Zheng and Bowen Yu and Chang Gao and Chengen Huang and Chenxu Lv and Chujie Zheng and Dayiheng Liu and Fan Zhou and Fei Huang and Feng Hu and Hao Ge and Haoran Wei and Huan Lin and Jialong Tang and Jian Yang and Jianhong Tu and Jianwei Zhang and Jianxin Yang and Jiaxi Yang and Jing Zhou and Jingren Zhou and Junyang Lin and Kai Dang and Keqin Bao and Kexin Yang and Le Yu and Lianghao Deng and Mei Li and Mingfeng Xue and Mingze Li and Pei Zhang and Peng Wang and Qin Zhu and Rui Men and Ruize Gao and Shixuan Liu and Shuang Luo and Tianhao Li and Tianyi Tang and Wenbiao Yin and Xingzhang Ren and Xinyu Wang and Xinyu Zhang and Xuancheng Ren and Yang Fan and Yang Su and Yichang Zhang and Yinger Zhang and Yu Wan and Yuqiong Liu and Zekun Wang and Zeyu Cui and Zhenru Zhang and Zhipeng Zhou and Zihan Qiu},
    journal = {arXiv preprint arXiv:2505.09388},
    year={2025}
}

@article{qwen2.5,
    title   = {Qwen2.5 Technical Report}, 
    author  = {An Yang and Baosong Yang and Beichen Zhang and Binyuan Hui and Bo Zheng and Bowen Yu and Chengyuan Li and Dayiheng Liu and Fei Huang and Haoran Wei and Huan Lin and Jian Yang and Jianhong Tu and Jianwei Zhang and Jianxin Yang and Jiaxi Yang and Jingren Zhou and Junyang Lin and Kai Dang and Keming Lu and Keqin Bao and Kexin Yang and Le Yu and Mei Li and Mingfeng Xue and Pei Zhang and Qin Zhu and Rui Men and Runji Lin and Tianhao Li and Tingyu Xia and Xingzhang Ren and Xuancheng Ren and Yang Fan and Yang Su and Yichang Zhang and Yu Wan and Yuqiong Liu and Zeyu Cui and Zhenru Zhang and Zihan Qiu},
    journal = {arXiv preprint arXiv:2412.15115},
    year    = {2024}
}

@inproceedings{Yao2025CacheBlend,
author = {Yao, Jiayi and Li, Hanchen and Liu, Yuhan and Ray, Siddhant and Cheng, Yihua and Zhang, Qizheng and Du, Kuntai and Lu, Shan and Jiang, Junchen},
title = {{CacheBlend}: Fast Large Language Model Serving for RAG with Cached Knowledge Fusion},
year = {2025},
booktitle = {Proceedings of the Twentieth European Conference on Computer Systems},
pages = {94–109},
numpages = {16},
}

@article{qwen2,
    title   = {Qwen2 Technical Report}, 
    author  = {An Yang and Baosong Yang and Binyuan Hui and Bo Zheng and Bowen Yu and Chang Zhou and Chengpeng Li and Chengyuan Li and Dayiheng Liu and Fei Huang and Guanting Dong and Haoran Wei and Huan Lin and Jialong Tang and Jialin Wang and Jian Yang and Jianhong Tu and Jianwei Zhang and Jianxin Ma and Jin Xu and Jingren Zhou and Jinze Bai and Jinzheng He and Junyang Lin and Kai Dang and Keming Lu and Keqin Chen and Kexin Yang and Mei Li and Mingfeng Xue and Na Ni and Pei Zhang and Peng Wang and Ru Peng and Rui Men and Ruize Gao and Runji Lin and Shijie Wang and Shuai Bai and Sinan Tan and Tianhang Zhu and Tianhao Li and Tianyu Liu and Wenbin Ge and Xiaodong Deng and Xiaohuan Zhou and Xingzhang Ren and Xinyu Zhang and Xipin Wei and Xuancheng Ren and Yang Fan and Yang Yao and Yichang Zhang and Yu Wan and Yunfei Chu and Yuqiong Liu and Zeyu Cui and Zhenru Zhang and Zhihao Fan},
    journal = {arXiv preprint arXiv:2407.10671},
    year    = {2024}
}

@article{zhao2025MPIC,
      title={{MPIC}: Position-Independent Multimodal Context Caching System for Efficient MLLM Serving}, 
      author={Shiju Zhao and Junhao Hu and Rongxiao Huang and Jiaqi Zheng and Guihai Chen},
      journal={arXiv preprint arXiv:2502.01960},
      year={2025},
}

@article{Yang2025KVShare,
      title={{KVShare}: An LLM Service System with Efficient and Effective Multi-Tenant KV Cache Reuse}, 
      author={Huan Yang and Renji Zhang and Mingzhe Huang and Weijun Wang and Yin Tang and Yuanchun Li and Yunxin Liu and Deyu Zhang},
      journal={arXiv preprint arXiv:2503.16525},
      year={2025},
}

@article{Qwen2.5-VL,
  title={Qwen2.5-VL Technical Report},
  author={Bai, Shuai and Chen, Keqin and Liu, Xuejing and Wang, Jialin and Ge, Wenbin and Song, Sibo and Dang, Kai and Wang, Peng and Wang, Shijie and Tang, Jun and Zhong, Humen and Zhu, Yuanzhi and Yang, Mingkun and Li, Zhaohai and Wan, Jianqiang and Wang, Pengfei and Ding, Wei and Fu, Zheren and Xu, Yiheng and Ye, Jiabo and Zhang, Xi and Xie, Tianbao and Cheng, Zesen and Zhang, Hang and Yang, Zhibo and Xu, Haiyang and Lin, Junyang},
  journal={arXiv preprint arXiv:2502.13923},
  year={2025}
}

@inproceedings{Dehghani2023Patch,
 author = {Dehghani, Mostafa and Mustafa, Basil and Djolonga, Josip and Heek, Jonathan and Minderer, Matthias and Caron, Mathilde and Steiner, Andreas and Puigcerver, Joan and Geirhos, Robert and Alabdulmohsin, Ibrahim M and Oliver, Avital and Padlewski, Piotr and Gritsenko, Alexey and Lucic, Mario and Houlsby, Neil},
 booktitle = {Advances in Neural Information Processing Systems},
 pages = {2252--2274},
 title = {Patch n’ Pack: NaViT, a Vision Transformer for any Aspect Ratio and Resolution},
 volume = {36},
 year = {2023}
}

@inproceedings{Zheng2024SGLang,
 author = {Zheng, Lianmin and Yin, Liangsheng and Xie, Zhiqiang and Sun, Chuyue and Huang, Jeff and Yu, Cody Hao and Cao, Shiyi and Kozyrakis, Christos and Stoica, Ion and Gonzalez, Joseph E. and Barrett, Clark and Sheng, Ying},
 booktitle = {Advances in Neural Information Processing Systems},
 pages = {62557--62583},
 title = {SGLang: Efficient Execution of Structured Language Model Programs},
 volume = {37},
 year = {2024}
}

@inproceedings{hu2025epic,
    title={{EPIC}: Efficient Position-Independent Caching for Serving Large Language Models},
    author={Junhao Hu and Wenrui Huang and Weidong Wang and Haoyi Wang and tiancheng hu and zhang qin and Hao Feng and Xusheng Chen and Yizhou Shan and Tao Xie},
    booktitle={Proceedings of the 42nd International Conference on Machine Learning},
    year={2025},
    }

@inproceedings{yue2025mmmupro,
    title = "{MMMU}-Pro: A More Robust Multi-discipline Multimodal Understanding Benchmark",
    author = "Yue, Xiang  and
      Zheng, Tianyu  and
      Ni, Yuansheng  and
      Wang, Yubo  and
      Zhang, Kai  and
      Tong, Shengbang  and
      Sun, Yuxuan  and
      Yu, Botao  and
      Zhang, Ge  and
      Sun, Huan  and
      Su, Yu  and
      Chen, Wenhu  and
      Neubig, Graham",
    booktitle = "Proceedings of the 63rd Annual Meeting of the Association for Computational Linguistics (Volume 1: Long Papers)",
    year = "2025",
    pages = "15134--15186",
}

@inproceedings{kwon2023efficient,
  title={Efficient Memory Management for Large Language Model Serving with PagedAttention},
  author={Woosuk Kwon and Zhuohan Li and Siyuan Zhuang and Ying Sheng and Lianmin Zheng and Cody Hao Yu and Joseph E. Gonzalez and Hao Zhang and Ion Stoica},
  booktitle={Proceedings of the ACM SIGOPS 29th Symposium on Operating Systems Principles},
  year={2023}
}

@article{Qwen2-VL,
  title={Qwen2-VL: Enhancing Vision-Language Model's Perception of the World at Any Resolution},
  author={Wang, Peng and Bai, Shuai and Tan, Sinan and Wang, Shijie and Fan, Zhihao and Bai, Jinze and Chen, Keqin and Liu, Xuejing and Wang, Jialin and Ge, Wenbin and Fan, Yang and Dang, Kai and Du, Mengfei and Ren, Xuancheng and Men, Rui and Liu, Dayiheng and Zhou, Chang and Zhou, Jingren and Lin, Junyang},
  journal={arXiv preprint arXiv:2409.12191},
  year={2024}
}

@article{Qwen-VL,
  title={Qwen-VL: A Versatile Vision-Language Model for Understanding, Localization, Text Reading, and Beyond},
  author={Bai, Jinze and Bai, Shuai and Yang, Shusheng and Wang, Shijie and Tan, Sinan and Wang, Peng and Lin, Junyang and Zhou, Chang and Zhou, Jingren},
  journal={arXiv preprint arXiv:2308.12966},
  year={2023}
}

@inproceedings{dosovitskiy2020vit,
  title={An Image is Worth 16x16 Words: Transformers for Image Recognition at Scale},
  author={Dosovitskiy, Alexey and Beyer, Lucas and Kolesnikov, Alexander and Weissenborn, Dirk and Zhai, Xiaohua and Unterthiner, Thomas and  Dehghani, Mostafa and Minderer, Matthias and Heigold, Georg and Gelly, Sylvain and Uszkoreit, Jakob and Houlsby, Neil},
  booktitle={International Conference on Learning Representations (ICLR)},
  year={2021}
}

@inproceedings{Vaswani2017Attention,
 author = {Vaswani, Ashish and Shazeer, Noam and Parmar, Niki and Uszkoreit, Jakob and Jones, Llion and Gomez, Aidan N and Kaiser, \L ukasz and Polosukhin, Illia},
 booktitle = {Advances in Neural Information Processing Systems 30},
 title = {Attention is all you need},
 volume = {30},
 year = {2017},
 pages = {5998--6008},
}

\newpage
\appendix
\section{Additional Results}

\subsection{Comparing Recomputing Earlier v.s. Later Tokens}

\begin{figure}[h]
    \centering
    \includegraphics[width=1.0\linewidth]{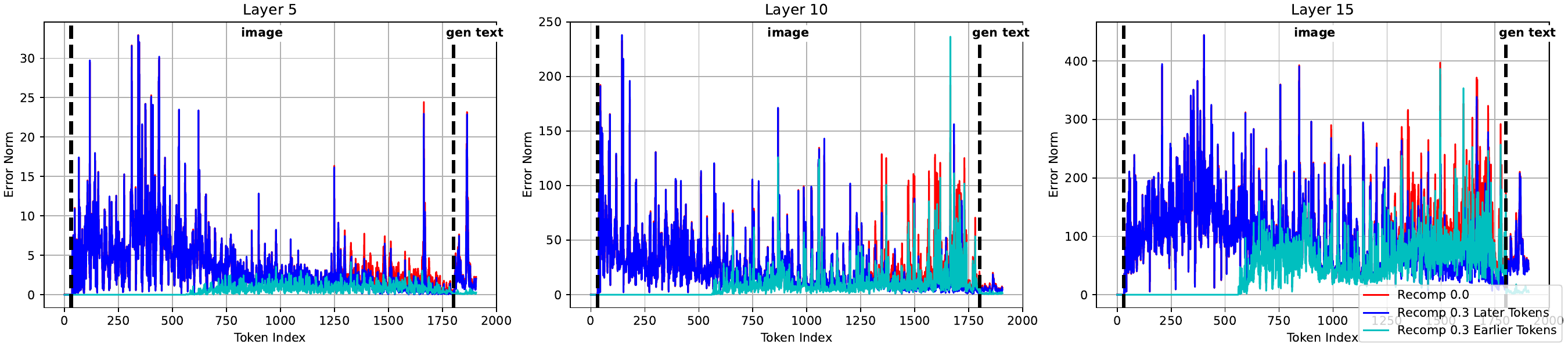}
    \caption{
        The reuse error of recomputing earlier tokens versus later tokens. Earlier token recomputation cancels out more reuse error and slows down the error propagation, leading to significantly lower error in generated texts.
    }
    \label{fig:attn_error_headtail}
\end{figure}

To determine which strategy yields lower reuse error, we compare allocating the recomputation budget to earlier versus later tokens. As shown in Fig.~\ref{fig:attn_error_headtail}, recomputing \textit{earlier tokens} significantly reduces the accumulated error. In contrast, recomputing later tokens provides only limited improvement, since early-stage errors have already propagated through the sequence, resulting in substantially higher overall error.

\subsection{Additional Results on Qwen2.5-VL Models}

To further demonstrate the effectiveness of our method, we also conducted relevant experiments on Qwen2.5VL-7B and Qwen2.5VL-32B. The speed and accuracy results are shown in the Table~\ref{tab:qwen2.5_vl_7b_ttft} - Table~\ref{tab:acc_2.5_32b} below. 

We can draw the same conclusion. As the recomputation rate increases from $r = 0.0$ toward $r = 1.0$, most datasets show a gradual upward trend. For instance, MMMU-VAL rises from 56.78 to 58.00, MathVision from 25.20 to 25.39, and RealWorldQA from 68.50 to 69.02. While a few datasets exhibit small oscillations (e.g., MMMU-Pro-Vision, AI2D-TEST), the overall average accuracy improves as more tokens are recomputed.

Moreover, the results highlight that the layer-wise dynamic KV cache recomputation scheme achieves higher accuracy than static recomputation at the same recomputation rate. This advantage is especially visible at intermediate rates (e.g., $r = 0.1 – 0.3$), where dynamic recomputation yields accuracy that is notably closer to the full-recompute values.

\subsection{Additional Results on Qwen3-VL Models}
To further demonstrate the effectiveness of our method, we evaluate the inference-time efficiency and accuracy trade-offs of various vision compression strategies on Qwen3-VL-30B-A3B model. 

As shown in Table~\ref{tab:qwen3_vl_30b_a3b_ttft}, though the overall acceleration is lower, we have consistently achieved a 20\%-80\% TTFT acceleration.
Crucially, Table~\ref{tab:acc_3_30a3b} demonstrates that aggressive compression incurs minimal accuracy degradation, i.e., VLCache with $\bar{r} = 0.035$ outperforms the baseline in mean accuracy (76.52 v.s. 76.47). This further supports the robustness of our approach. Overall, these results reveal a highly favorable Pareto frontier, i.e., up to 1.79× faster TTFT with no loss, and sometimes improvement in accuracy.

\begin{table}[t]
\centering
\caption{Speedup for Qwen2.5-VL-7B}
\label{tab:qwen2.5_vl_7b_ttft}
\resizebox{\textwidth}{!}{
\begin{tabular}{lcccccccccc}
\toprule
        & \multicolumn{2}{c}{\textbf{1.4K}} & \multicolumn{2}{c}{\textbf{3.6K}} & \multicolumn{2}{c}{\textbf{5K}} & \multicolumn{2}{c}{\textbf{10K}} & \multicolumn{2}{c}{\textbf{20K}} \\
        \cmidrule(lr){2-3}\cmidrule(lr){4-5}\cmidrule(lr){6-7}\cmidrule(lr){8-9}\cmidrule(lr){10-11}
        \textbf{Configuration} & \textbf{TTFT} & \textbf{Speedup} & \textbf{TTFT} & \textbf{Speedup} & \textbf{TTFT} & \textbf{Speedup} & \textbf{TTFT} & \textbf{Speedup} & \textbf{TTFT} & \textbf{Speedup} \\
\midrule
        Origin                                         & 0.635 & 1.00x & 2.631 & 1.00x & 4.398 & 1.00x & 16.160 & 1.00x & 62.750 & 1.00x \\
        w/o ViT                                       & 0.322 & 1.97x & 0.985 & 2.67x & 1.334 & 3.30x & 2.693  & 6.00x & 6.251  & 10.04x \\
        Static (r=0.3)            & 0.251 & 2.53x & 0.722 & 3.64x & 0.987 & 4.46x & 2.006  & 8.06x & 4.402  & 14.25x \\
        Static (r=0.2)            & 0.231 & 2.75x & 0.673 & 3.91x & 0.936 & 4.70x & 1.877  & 8.61x & 4.054  & 15.48x \\
        Static (r=0.1)            & 0.216 & 2.94x & 0.612 & 4.30x & 0.844 & 5.21x & 1.740  & 9.29x & 3.847  & 16.31x \\
        Static (r=0.0)            & 0.210 & 3.02x & 0.606 & 4.34x & 0.825 & 5.33x & 1.626  & 9.94x & 3.701  & 16.95x \\
        Dynamic ($\bar{r}=0.1$)   & 0.255 & 2.49x & 0.712 & 3.70x & 0.957 & 4.60x & 1.912  & 8.45x & 4.130  & 15.19x \\
        Dynamic ($\bar{r}=0.05$)  & 0.232 & 2.74x & 0.699 & 3.76x & 0.879 & 5.00x & 1.885  & 8.57x & 3.966  & 15.82x \\
\bottomrule
\end{tabular}
}
\end{table}

\begin{table}[htbp]
\centering
\caption{Accuracy Preservation for Qwen2.5-VL-7B}
\label{tab:acc_2.5_7b}
\resizebox{\textwidth}{!}{
\begin{tabular}{lccccccccc}
\toprule
\textbf{Dataset} &
$r=0.0$ &
$\bar{r}=0.01$ &
$\bar{r}=0.02$ &
$\bar{r}=0.05$ &
$r=0.1$ &
$\bar{r}=0.1$ &
$r=0.2$ &
$r=0.3$ &
$r=1.0$ \\
\midrule
MMMU-VAL             & 56.78 & 56.78 & 57.11 & 57.78 & 57.00 & 57.33 & 57.33 & 57.44 & \textbf{58.00} \\
MMMU-Pro-Standard   & 41.16 & 41.50 & \textbf{41.56} & \textbf{41.56} & 40.87 & 41.10 & 41.10 & 40.69 & 40.87 \\
MMMU-Pro-Vision     & 33.24 & 34.45 & 34.91 & 34.91 & 34.68 & \textbf{35.90} & 35.09 & 36.99 & \textbf{36.99} \\
MathVista-MINI       & 67.00 & 67.20 & 65.80 & 66.90 & 67.40 & 67.60 & \textbf{68.80} & 68.50 & 68.40 \\
MathVision            & 25.20 & 24.51 & 25.66 & 26.05 & 25.82 & \textbf{26.35} & 25.56 & 26.22 & 25.39 \\
HallusionBench        & 50.54 & 50.46 & 51.24 & 50.37 & 50.15 & 50.66 & 50.76 & 50.81 & \textbf{53.13} \\
MBench-DEV-CN-V11  & 80.88 & 81.04 & 80.88 & 80.96 & 81.04 & 80.88 & 81.11 & 81.35 & \textbf{82.20} \\
MBench-DEV-EN-V11  & 83.20 & 83.13 & 83.28 & 83.13 & 83.05 & 82.89 & 82.89 & 83.13 & \textbf{84.13} \\
RealWorldQA           & 68.50 & 68.50 & 68.76 & 68.37 & 68.50 & 68.63 & 68.24 & 68.50 & \textbf{69.02} \\
AI2D-TEST            & 84.26 & 84.13 & 84.23 & \textbf{84.26} & 84.07 & 84.20 & 84.03 & 84.20 & 84.13 \\
ChartQA-TEST         & 86.08 & 85.92 & 86.24 & 86.20 & 86.40 & \textbf{86.44} & 86.24 & 86.40 & 87.40 \\
DocVQA-VAL           & 94.61 & 94.61 & 94.60 & 94.58 & 94.60 & 94.63 & 94.58 & 94.43 & \textbf{94.74} \\
\midrule
\textbf{Mean}        & 64.29 & 64.35 & 64.52 & 64.59 & 64.47 & 64.72 & 64.64 & 64.88 & \textbf{65.37} \\
\bottomrule
\end{tabular}
}
\end{table}

\begin{table}[htbp]
\centering
\caption{Speedup for Qwen2.5-VL-32B}
\label{tab:qwen2.5_vl_32b_ttft}
\resizebox{\textwidth}{!}{
\begin{tabular}{lcccccccccc}
\toprule
        & \multicolumn{2}{c}{\textbf{1.4K}} & \multicolumn{2}{c}{\textbf{3.6K}} & \multicolumn{2}{c}{\textbf{5K}} & \multicolumn{2}{c}{\textbf{10K}} & \multicolumn{2}{c}{\textbf{20K}} \\
        \cmidrule(lr){2-3}\cmidrule(lr){4-5}\cmidrule(lr){6-7}\cmidrule(lr){8-9}\cmidrule(lr){10-11}
        \textbf{Configuration} & \textbf{TTFT} & \textbf{Speedup} & \textbf{TTFT} & \textbf{Speedup} & \textbf{TTFT} & \textbf{Speedup} & \textbf{TTFT} & \textbf{Speedup} & \textbf{TTFT} & \textbf{Speedup} \\
\midrule
        Origin                                & 1.176 & 1.00x & 3.961 & 1.00x & 6.525 & 1.00x & 20.110 & 1.00x & 70.570 & 1.00x \\
        w/o ViT                              & 0.873 & 1.35x & 2.342 & 1.69x & 3.226 & 2.02x & 6.865 & 2.93x & 15.351 & 4.60x \\
        Static (r=0.3)             & 0.503 & 2.34x & 1.303 & 3.04x & 1.777 & 3.67x & 3.494 & 5.76x & 7.649 & 9.23x \\
        Static (r=0.2)             & 0.457 & 2.57x & 1.073 & 3.69x & 1.488 & 4.39x & 2.969 & 6.77x & 6.308 & 11.19x \\
        Static (r=0.1)             & 0.375 & 3.14x & 0.944 & 4.20x & 1.275 & 5.12x & 2.707 & 7.43x & 5.444 & 12.96x \\
        Static (r=0.0)             & 0.325 & 3.62x & 0.620 & 6.39x & 0.820 & 7.96x & 1.905 & 10.56x & 4.181 & 16.88x \\
        Dynamic ($\bar{r}=0.04$)   & 0.391 & 3.01x & 0.929 & 4.26x & 1.181 & 5.52x & 2.443 & 8.23x & 5.406 & 13.05x \\
        Dynamic ($\bar{r}=0.02$)   & 0.397 & 2.96x & 0.920 & 4.31x & 1.201 & 5.43x & 2.469 & 8.14x & 5.286 & 13.35x \\
\bottomrule
\end{tabular}
}
\end{table}

\begin{table}[t]
\centering
\caption{Accuracy Preservation for Qwen2.5-VL-32B}
\label{tab:acc_2.5_32b}
\resizebox{\textwidth}{!}{
\begin{tabular}{lccccccc}
\toprule
\textbf{Dataset} &
$r=0.0$ &
$\bar{r}=0.02$ &
$\bar{r}=0.04$ &
$r=0.1$ &
$r=0.2$ &
$r=0.3$ &
$r=1.0$ \\
\midrule
MMMU-VAL             & 66.56 & 67.00 & \textbf{67.89} & \textbf{67.89} & 66.67 & 66.78 & 67.00 \\
MMMU-Pro-Standard   & 51.50 & \textbf{52.25} & 52.08 & 51.85 & 51.91 & 51.62 & 51.91 \\
MMMU-Pro-Vision     & 45.26 & 46.65 & \textbf{47.28} & 45.95 & 45.78 & 46.53 & 46.59 \\
MathVista-MINI       & \textbf{75.70} & 75.30 & 75.60 & 75.00 & 74.20 & 74.10 & 74.30 \\
MathVision            & 38.55 & 38.95 & 38.45 & 39.61 & \textbf{39.87} & 39.77 & 39.41 \\
HallusionBench        & 57.20 & 56.78 & 57.21 & \textbf{57.55} & 57.46 & 56.41 & 57.17 \\
MBench-DEV-CN-V11  & 85.14 & \textbf{85.53} & 85.29 & 85.22 & 85.06 & 85.22 & 85.45 \\
MBench-DEV-EN-V11  & 86.30 & 86.46 & 86.53 & 86.22 & \textbf{86.84} & 86.76 & 86.61 \\
RealWorldQA           & 69.54 & 70.20 & 70.46 & 69.54 & 69.67 & 69.41 & \textbf{70.85} \\
AI2D-TEST            & 85.33 & 85.65 & 85.30 & 85.78 & 85.85 & \textbf{85.88} & 85.52 \\
ChartQA-TEST         & 68.92 & 69.72 & 69.44 & 69.64 & 70.20 & 70.92 & \textbf{74.60} \\
DocVQA-VAL           & 92.52 & 92.55 & 92.54 & \textbf{92.73} & 92.44 & 92.50 & 92.68 \\
\midrule
\textbf{Mean}        & 68.54 & 68.92 & 69.01 & 68.92 & 68.83 & 68.83 & \textbf{69.34} \\
\bottomrule
\end{tabular}
}
\end{table}

\begin{table}[t]
\centering
\caption{Speedup for Qwen3-VL-30B-A3B}
\label{tab:qwen3_vl_30b_a3b_ttft}
\resizebox{\textwidth}{!}{
\begin{tabular}{lcccccccccc}
\toprule
        & \multicolumn{2}{c}{\textbf{1K}} & \multicolumn{2}{c}{\textbf{3K}} & \multicolumn{2}{c}{\textbf{5K}} & \multicolumn{2}{c}{\textbf{10K}} & \multicolumn{2}{c}{\textbf{20K}} \\
        \cmidrule(lr){2-3}\cmidrule(lr){4-5}\cmidrule(lr){6-7}\cmidrule(lr){8-9}\cmidrule(lr){10-11}
        \textbf{Configuration} & \textbf{TTFT} & \textbf{Speedup} & \textbf{TTFT} & \textbf{Speedup} & \textbf{TTFT} & \textbf{Speedup} & \textbf{TTFT} & \textbf{Speedup} & \textbf{TTFT} & \textbf{Speedup} \\
\midrule
        Origin                     & 0.350 & 1.00x & 1.052 & 1.00x & 1.802 & 1.00x & 3.758 & 1.00x & 8.172 & 1.09x \\
        w/o ViT                    & 0.322 & 1.09x & 0.815 & 1.29x & 1.418 & 1.27x & 3.006 & 1.25x & 6.931 & 1.18x \\
        Static (r=0.3)             & 0.314 & 1.11x & 0.756 & 1.39x & 1.233 & 1.46x & 2.447 & 1.54x & 4.997 & 1.64x \\
        Static (r=0.2)             & 0.299 & 1.17x & 0.750 & 1.40x & 1.245 & 1.45x & 2.369 & 1.59x & 4.804 & 1.70x \\
        Static (r=0.1)             & 0.283 & 1.24x & 0.752 & 1.40x & 1.280 & 1.41x & 2.462 & 1.53x & 4.736 & 1.73x \\
        Static (r=0.0)             & 0.327 & 1.07x & 0.747 & 1.41x & 1.226 & 1.76x & 2.385 & 1.58x & 4.697 & 1.74x \\
        Dynamic ($\bar{r}=0.035$)  & 0.310 & 1.13x & 0.761 & 1.38x & 1.215 & 1.47x & 2.325 & 1.62x & 4.562 & 1.79x \\
        Dynamic ($\bar{r}=0.025$)  & 0.317 & 1.10x & 0.785 & 1.34x & 1.216 & 1.48x & 2.311 & 1.63x & 4.572 & 1.79x \\
\bottomrule
\end{tabular}
}
\end{table}

\begin{table}[t]
\centering
\caption{Accuracy Preservation for Qwen3-VL-30B-A3B}
\label{tab:acc_3_30a3b}
\resizebox{\textwidth}{!}{
\begin{tabular}{lccccccc}
\toprule
\textbf{Dataset} &
$r=0.0$ &
$\bar{r}=0.025$ &
$\bar{r}=0.035$ &
$r=0.1$ &
$r=0.2$ &
$r=0.3$ &
$r=1.0$ \\
\midrule
MMMU-VAL            & 71.00 & 69.33 & 70.00 & 69.56 & 69.44 & 70.56 & \textbf{70.89} \\
MMMU-Pro-Standard   & 59.83 & 60.75 & 60.06 & 61.27 & \textbf{61.56} & 60.40 & 60.46 \\
MMMU-Pro-Vision     & 53.35 & 53.53 & 54.51 & 54.05 & 53.35 & 54.57 & \textbf{55.49} \\
MathVista-MINI      & 78.10 & 79.80 & \textbf{80.10} & 79.70 & 79.60 & 79.90 & 78.20 \\
HallusionBench      & 60.61 & 60.75 & 61.07 & 60.75 & 60.65 & 61.80 & \textbf{62.20} \\
MBench-DEV-CN-V11   & 87.15 & 86.53 & \textbf{87.31} & 87.00  & 86.61 & 87.23 & 86.76 \\
MBench-DEV-EN-V11   & 87.38 & 87.62 & 88.24 & 87.54 & 87.85 & 88.08 & \textbf{88.47} \\
RealWorldQA         & 74.12 & 73.99 & \textbf{75.16} & 74.51  & 73.99 & 73.73 & 73.33 \\
AI2D-TEST           & 85.40 & \textbf{85.98} & 85.04 & 85.69 & 85.40 & 85.49 & 84.81 \\
ChartQA-TEST        & 85.60 & 85.76 & 85.72 & 85.60 & 85.84 & \textbf{86.04} & 85.96 \\
DocVQA-VAL          & 94.47 & 94.36 & 94.48 & 94.32 & 94.36 & 94.41 & \textbf{94.59} \\
\midrule
\textbf{Mean}        & 76.09 & 76.22 & 76.52 & 76.36 & 76.24 & \textbf{76.56} & 76.47 \\
\bottomrule
\end{tabular}
}
\end{table}

\end{document}